\documentclass[10pt,journal,compsoc]{IEEEtran}
\usepackage{epsfig}
\usepackage{graphicx}
\usepackage{amsmath}
\usepackage{amssymb}
\usepackage{wasysym}
\usepackage{tabularx}
\usepackage{xcolor}
\usepackage{threeparttable}
\usepackage{algorithm}
\usepackage{algorithmic}
\usepackage{colortbl}
\usepackage{multirow}
\usepackage{graphicx}
\usepackage{subfig}
\usepackage{ragged2e}
\usepackage{bm}
\usepackage[pagebackref=false,breaklinks=true,colorlinks,bookmarks=false]{hyperref}
\usepackage{caption}
\ifCLASSOPTIONcompsoc
  \usepackage[nocompress]{cite}
\else
  \usepackage{cite}
\fi
\ifCLASSINFOpdf
\else
\fi

\usepackage{xspace}
\makeatletter
\DeclareRobustCommand\onedot{\futurelet\@let@token\@onedot}
\def\@onedot{\ifx\@let@token.\else.\null\fi\xspace}

\newcolumntype{P}[1]{>{\centering\arraybackslash}p{#1}}
\begin{document}

\title{Exploring Effective Priors and Efficient Models for Weakly-Supervised Change Detection}

\author{Zhenghui Zhao, Lixiang Ru, Chen Wu

%~\IEEEmembership{Staff,~IEEE,}
        % <-this % stops a space
\IEEEcompsocitemizethanks{
\IEEEcompsocthanksitem This work is partly supported by the National Key Research and Development Program of China under Grant 2022YFB3903300, and partly by the National Natural Science Foundation of China under Grant T2122014, 61971317. (\textit{Corresponding
author: Chen Wu}.)
\IEEEcompsocthanksitem Z. Zhao and C. Wu are with the State Key Laboratory of Information Engineering in Surveying, Mapping and Remote Sensing, and Institute of Artificial Intelligence, Wuhan University, Wuhan 430072, China. 

\IEEEcompsocthanksitem L. Ru is with Ant Group, Hangzhou 310013, China. 
%% E-mail: zhaozhenghui@whu.edu.cn; chen.wu@whu.edu.cn.
}}

\markboth{}%
{Shell \MakeLowercase{\textit{et al.}}: Bare Demo of IEEEtran.cls for Computer Society Journals}

\IEEEtitleabstractindextext{%
\begin{abstract}
\justifying
Weakly-supervised change detection (WSCD) aims to detect pixel-level changes with only image-level annotations. Owing to its label efficiency, WSCD is drawing increasing attention recently. However, current WSCD methods often encounter the challenge of change missing and fabricating, i.e., the inconsistency between image-level annotations and pixel-level predictions. Specifically, change missing refer to the situation that the WSCD model fails to predict any changed pixels, even though the image-level label indicates changed, and vice versa for change fabricating. To address this challenge, in this work, we leverage global-scale and local-scale priors in WSCD and propose two components: a Dilated Prior (DP) decoder and a Label Gated (LG) constraint. The DP decoder decodes samples with the changed image-level label, skips samples with the unchanged label, and replaces them with an all-unchanged pixel-level label. The LG constraint is derived from the correspondence between changed representations and image-level labels, penalizing the model when it mispredicts the change status. Additionally, we develop TransWCD, a simple yet powerful transformer-based model, showcasing the potential of weakly-supervised learning in change detection. By integrating the DP decoder and LG constraint into TransWCD, we form TransWCD-DL. Our proposed TransWCD and TransWCD-DL achieve significant +6.33\% and +9.55\% F1 score improvements over the state-of-the-art methods on the WHU-CD dataset, respectively. Some performance metrics even exceed several fully-supervised change detection (FSCD) competitors. Code will be available at \url{https://github.com/zhenghuizhao/TransWCD}.
\end{abstract}
\begin{IEEEkeywords}
Remote Sensing, Change Detection, Weakly-Supervised Learning, Prior knowledge, Informed Learning
\end{IEEEkeywords}}

% make the title area
\maketitle

\IEEEdisplaynontitleabstractindextext

\IEEEpeerreviewmaketitle

\begin{figure}
	\centering
	\includegraphics[scale=0.35]{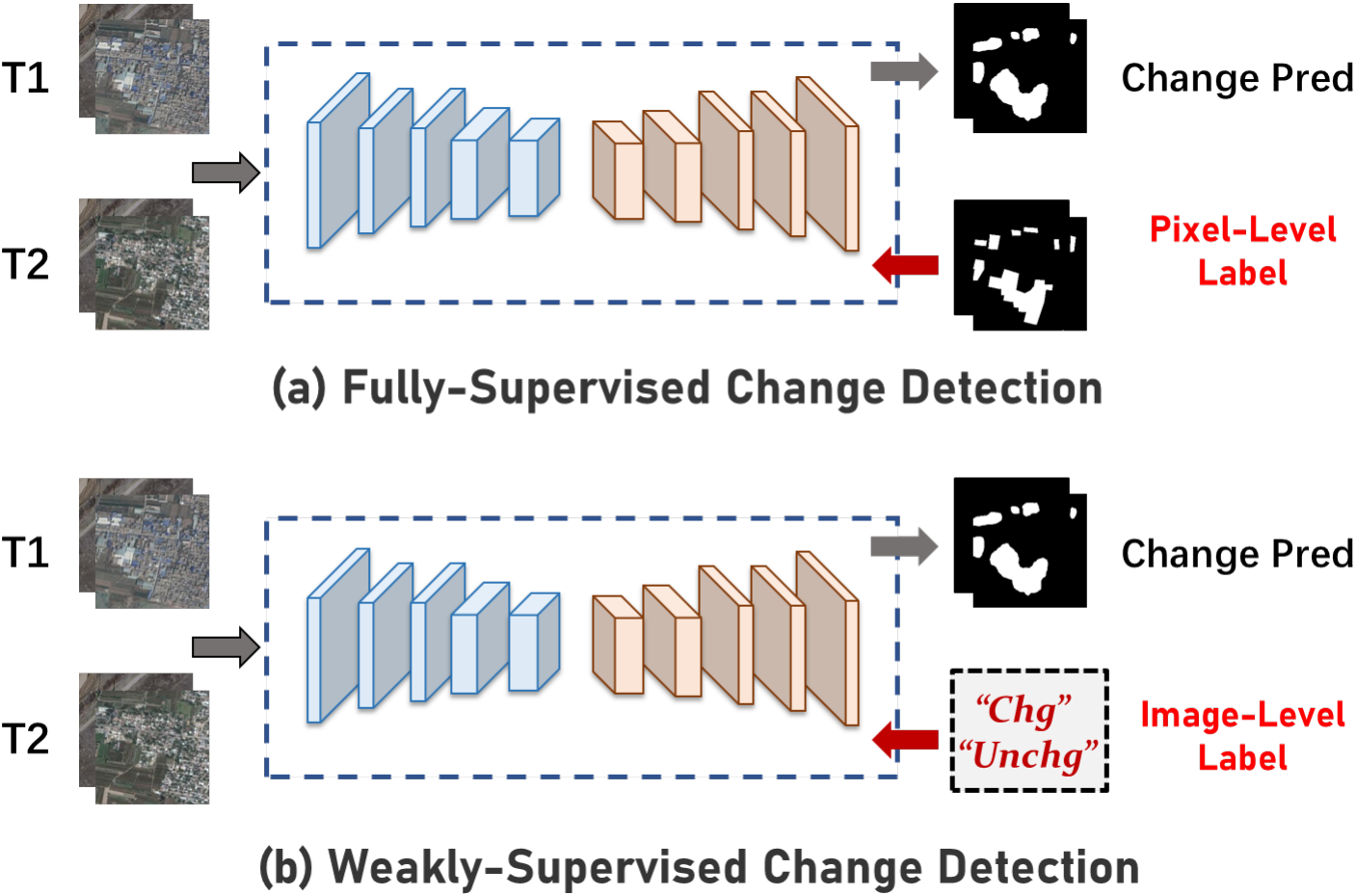}
	\caption{Difference between fully-supervised change detection (FSCD) and weakly-supervised change detection (WSCD). WSCD utilizes image-level binary labels "\textit{Changed (Chg)}" and "\textit{Unchanged (Unchg)}"  as supervision signals instead of ground truths. }
	\label{introduction}
\end{figure}

\IEEEraisesectionheading{\section{Introduction}\label{sec:intro}}
\IEEEPARstart{C}hange detection (CD) is a fundamental task in remote sensing that determines landscape changes over time within a given geographical area. This task assigns a binary label to each pixel by comparing coregistered multi-temporal images. CD has a wide range of applications, including disaster prevention \cite{11wu}, damage assessment \cite{xu2019building}, land-use surveying \cite{7wu, 8wu}, urban development \cite{lee2021local}, and environmental monitoring \cite{9wu, song2014remote}. Recent advancements in deep learning have sparked significant breakthroughs in CD \cite{21wu}. However, most existing CD tasks heavily rely on fully-supervised approaches that necessitate pixel-level annotations \cite{40wu, 41wu}, rendering them time-consuming and labor-intensive. With the proliferation of high-resolution satellites launched worldwide, there has been growing interest in weakly-supervised change detection (WSCD) \cite{Khan2017, ander2020, kal2021, wu2023, Huang2023}. Fig. \ref{introduction} illustrates the comparison between fully-supervised change detection (FSCD) and WSCD. WSCD leverages more affordable image-level labels as supervision signals instead of pixel-level ground truths, emerging a promising alternative with improved performance-cost tradeoffs. Pioneering works \cite{Khan2017, ander2020, kal2021, wu2023, Huang2023} have recognized weak supervision's benefits and proposed outstanding WSCD methods. 

\begin{figure}
	\centering
	\includegraphics[scale=0.27]{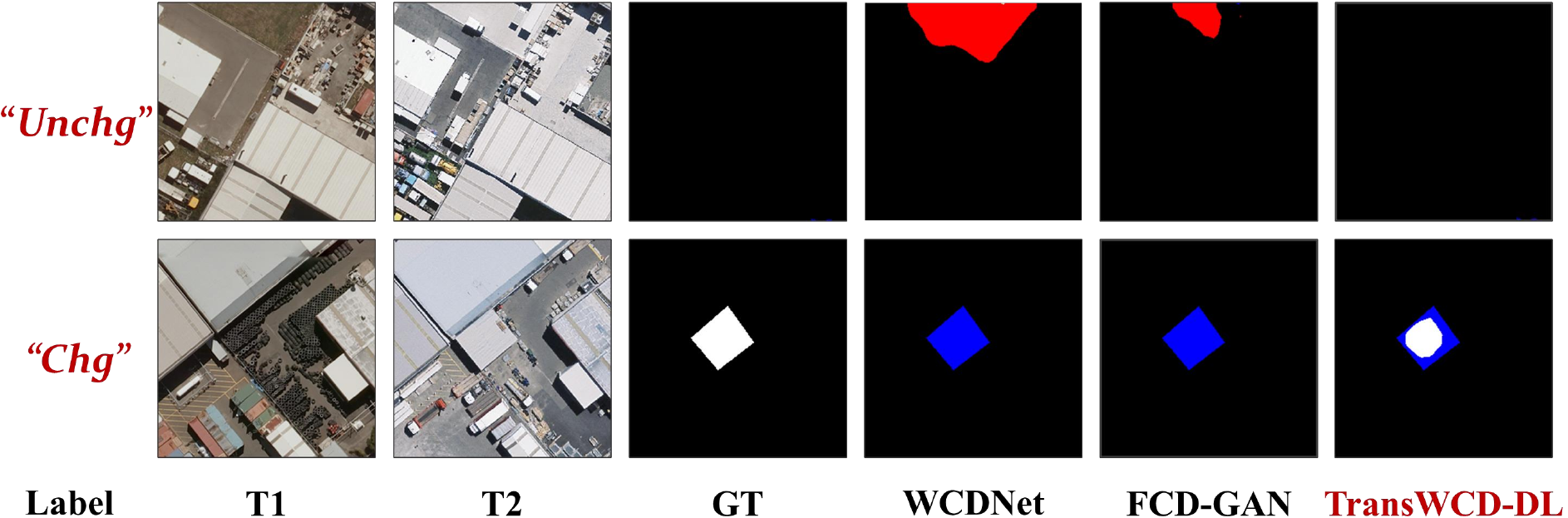}
	\caption{Problem of change missing and fabricating in WSCD.  \textcolor{red}{False positives} and \textcolor{blue}{false negatives} are highlighted in red and blue, respectively. The 1st row illustrates the phenomenon of change fabricating in WCDNet \cite{ander2020} and FCD-GAN \cite{wu2023}: the image-level label suggests no changes, but changed pixels are incorrectly identified. The 2nd row showcases change missing: the image-level label indicates changes, but no corresponding changed pixels are detected. Our TransWCD-DL significantly improves change missing and fabricating. For a comparison with more existing methods, refer to Fig. \ref{experiment3}.}
	\label{introduction2}
\end{figure}

The inception of WSCD can be traced back to 2017 \cite{Khan2017}. Since then, only a limited number of  publications have been dedicated to WSCD, including \cite{ander2020, kal2021, wu2023, Huang2023}. These studies employ diverse techniques, with Khan \textit{et al.} \cite{Khan2017} and Andermatt \textit{et al.} \cite{ander2020} utilizing Conditional Random Fields (CRF), Kalita \textit{et al.} \cite{kal2021} applying classic Principal Component Analysis (PCA) in combination with the K-means clustering algorithm, Wu \textit{et al.} \cite{wu2023} employing Generative Adversarial Networks (GAN), and Huang \textit{et al.} \cite{Huang2023} introducing a novel method of sample augmentation. While these methods contribute significantly to the advancement of WSCD, they overlook a crucial challenge: change missing and fabricating. As depicted in Fig. \ref{introduction2}, existing works encounter scenarios where the image-level label indicate the presence of changes, but the model fails to predict any changed pixels (i.e., change missing). Additionally, there are cases that the image-level label suggests no changes, yet the model predicts changed pixels (i.e., change fabricating). The variability of environmental conditions and the unavailability of ground truths exacerbate the inconsistency problem between image-level annotations and pixel-level predictions. 

Inspired by informed learning \cite{2021informed}, We incorporate global-scale and local-scale prior knowledge underlying WSCD into the model to guide precise learning and address the challenge of changed missing and fabricating. The specific prior knowledge observed is as follows: 1) \textit{The global-scale prior states that an unchanged image-level label corresponds to no change, aligning with an unchanged pixel-level ground truth.} 2) \textit{The local-scale prior indicates that a change in the image-level label implies the existence of changed regions; conversely, if the image-level label is unchanged, there are no changed regions.}

Based on these observations for WSCD, we design the corresponding Dilated Prior (DP) decoder and Label Gated (LG) constraint. The DP decoder incorporates the global-scale prior through the utilization of a dilated convolution group, applied to the final-stage feature map, competently addressing change fabricating. In situations where the image-level label remains unchanged, DP decoder refrains from decoding the input from the model upstream. Instead, it employs an all-unchanged pixel-level label as the supervision signal. LG constraint, combining the local-scale prior, serves as a feature-level penalty mechanism by the unit impulse function. It is explicitly activated in worst-case scenarios where the translation from image-level labels to pixel-level predictions fails, reducing the occurrence of change missing and fabricating, resulting in improved performance.

Moreover, drawing insights from weakly supervised approaches in other visual tasks \cite{zhang2021weakly, minaee2021image, ru2022}, these existing works may only harness the partial performance potential of weakly supervised paradigm in CD. Hence, we undertake further exploration to develop an efficient WSCD model. Unlike previous WSCD methods based on Convolutional Neural Networks (ConvNets), we propose TransWCD, a simple but strong transformer-based model tailored for WSCD, competently capturing more global dependencies  \cite{gao2021ts, wu2020visual, zhang2020feature}. Specifically, our TransWCD contains three constituent modules: 1) hierarchical transformer encoder with image-level supervision, 2) mono-layer minimalist difference, and 3) shortcut-free multi-scale Class Activation Map (CAM) prediction. These modules form two architectures: single-stream and Siamese dual-stream. TransWCD adopts an encoder-only architecture, and merges the DP decoder and LG constraint, culminating in TransWCD-DL.

Despite its unassuming appearance, TransWCD exhibits exceptional performance surpassing the state-of-the-art \cite{Huang2023} on WHU-CD dataset, highlighting the remarkable potential of weakly-supervised learning in CD. TransWCD effectively bridges the performance gap between FSCD and WSCD, establishing itself as a compelling baseline for future WSCD algorithms. As such, TransWCD is non-trivial to the CD field. Furthermore, Our TransWCD-DL incorporates prior knowledge, leading to further improvements in change predictions, addressing the challenge of changed missing and fabricating, and enhancing model interpretability. In addition, our research uncovers the underlying factors that contribute to the exceptional performance of CAMs in WSCD prediction. This analysis provides valuable insights into WSCD, shedding light on the mechanisms behind CAMs' effectiveness. We also delve into the necessity of additional shortcut connections in CAM prediction for transformer-based WSCD. 

In summary, our contributions are as follows:
\begin{itemize}
\item{We introduce the global-scale and local-scale priors in weakly-supervised change detection (WSCD) and leverage them to design a Dilated Prior (DP) decoder and a Label Gated (LG) constraint. These components competently tackle the dilemma of changed missing and fabricating.}
\item{We develop TransWCD, an exceptionally efficient model tailored for WSCD, which could serve as an optional baseline for future works, demonstrating the true potential of weak supervision in CD.}
\item{Our study represents the first comprehensive exploration of applying transformers to WSCD, despite their longstanding popularity. We conduct extensive discussions on transformer-based WSCD that offer promising opportunities for advancing the WSCD field.}
\end{itemize}

\section{Related Work}\label{sec:related}
\subsection{Transformer for Dense Prediction}
Transformers have recently emerged as powerful vision architectures, owing to the ability to capture global context and long-range dependencies. The first pure transformer architecture for computer vision is ViT \cite{dos2020image}. Subsequently, DeiT \cite{tou2021} improves the training on small datasets with effective training strategies, making it useful for vision tasks in situations lacking significant amounts of training data. Swin Transformer \cite{liu2021swin} reduces the computational complexity by utilizing shifted windowing to calculate self-attention within a local window. These endeavors and others led to the gradual replacement of ConvNets by transformers as the dominant neural network architecture in visual tasks.

As transformers have gained momentum, several backbone architectures tailored for pixel-level dense prediction tasks have emerged. SETR \cite{zheng2021} treats semantic segmentation as a sequence-to-sequence prediction and designs a transformer-based encoder with three types of decoders. PVT \cite{2021pyramid} adopts a pyramid encoder to capture multi-scale representations, achieving notable performance in semantic segmentation and object detection. Segformer \cite{segformer} comprises five multi-stage encoder constructions and an all-MLP decoder for semantic segmentation. Additionally, it eliminates position embedding and incorporates a grouped attention mechanism to improve computation. 

\subsection{Change Detection}
{\bf{Fully-supervised change detection.}} Most CD methods in recent years have relied on deep learning, focusing predominantly on pixel-level fully supervised approaches. Early methods utilized ConvNet-based architectures, while transformer-based methods have recently gained popularity. Daudt \textit{et al.} \cite{47daudt2018fully} design the first end-to-end CD method with the U-Net backbone, containing three effective architectures FC-EF, FC-Siam-conc, and FC-Siam-diff. Subsequently, in addition to exploiting U-Net, many works use VGG16 \cite{simonyan2014very} and the ResNet series \cite{he2016deep} as their backbones. For instance, IFNet \cite{35zhang2020deeply} adopts VGG16 as the backbone; DTCDSCN \cite{40liu2021building} and DMINet \cite{feng2023} employ ResNet34 and ResNet18 as their backbones, respectively. In the case of transformer-based works, SwinSUNet \cite{Zhang2022} and MSTDSNet-CD \cite{song2022} adopt Swin Transformer \cite{liu2021swin} as their backbone, and BIT \cite{31chen2021remote} combines ResNet18 with ViT. 

These noteworthy methods primarily involve multi-level representation learning to achieve optimal performance. For instance, DMINet \cite{feng2023} even achieves a 90.71\% F1 score under the general settings of the LEVIR-CD dataset, demonstrating that multi-scale features are crucial to CD. However, resorting to labor-intensive fully-supervised learning for large-scale remote sensing data could be impractical and inefficient.

{\bf{Weakly-supervised change detection.}} Some researchers have recognized the limitations of fully-supervised learning and have conducted WSCD studies using image-level annotation. Khan \textit{et al.} \cite{Khan2017} propose the first WSCD work under deep learning. WCDNet \cite{ander2020} designs a remapping block and refines the change mask with CRF-RNN. Kalita \textit{et al.}  \cite{kal2021} combine the PCA and K-means algorithm with a custom Siamese convolutional network. BGMix \cite{Huang2023} proposes a background-mixed augmentation, cooperating with an augmented \& real data consistency loss. FCD-GAN \cite{wu2023} develops a universal model with multiple supervised learning for CD, based on the generative adversarial network with fully convolutional networks, involving weakly supervised learning. These advancements have significantly propelled the progress of WSCD, but overlooked the issue of change missing and fabricating, pronounced and exacerbated by the absence of ground truths.

Additionally, Aside from Khan \textit{et al.} \cite{Khan2017} utilizing a self-designed convolutional network, the remaining works \cite{ander2020,kal2021,Huang2023,wu2023} exploit VGG16 as their backbone, and the performance of Huang \textit{et al.} \cite{Huang2023} is achieved by utilizing VGG16 models. All of these previous works are ConvNets-based. WSCD, under the currently popular transformers, is left to explore. 

\subsection{Informed Machine Learning}
Despite its great success, there are many circumstances where purely data-driven approaches may reach their limits for machine learning, mainly when dealing with insufficient data. Moreover, purely data-driven models might not adhere to constraints imposed by natural laws, regulatory guidelines, or security protocols, which are fundamental for establishing trustworthiness in AI \cite{2020toward}. By incorporating prior knowledge into the learning process, also known as informed machine learning \cite{2021informed}, these issues can be skillfully mitigated, simultaneously reducing computational overhead and enhancing the interpretability of models \cite{2019explainable}.

This prior knowledge encompasses a range of forms, including specialized and formalized scientific knowledge, general world knowledge, and intuitive expert knowledge \cite{2021informed}. The integration of prior knowledge into the models can be achieved through diverse approaches, such as algebraic equations \cite{2003knowledgebased}, differential equations \cite{2017physical}, simulation results \cite{shrivastava2017}, logic rules \cite{hu2016harnessing}, spatial invariances \cite{bronstein2017}, probabilistic relations \cite{angelopoulos2008}, knowledge graphs \cite{bian2014knowledge}, and human feedback \cite{heckerman1995}. 

In this paper, we leverage logic rules to incorporate our observations of the global-scale prior to WSCD, forming DP decoder. Additionally, leveraging algebraic equations, we seamlessly integrate the local-scale prior to WSCD into the learning objective function by imposing LG constraint. These two modules enable the model to learn more logically, alleviating the problems of change missing and fabricating in WSCD.

%-------------------------------------------------------------------------
\section{Methodology}\label{sec:method}
\begin{figure}
	\centering
	\includegraphics[scale=0.46]{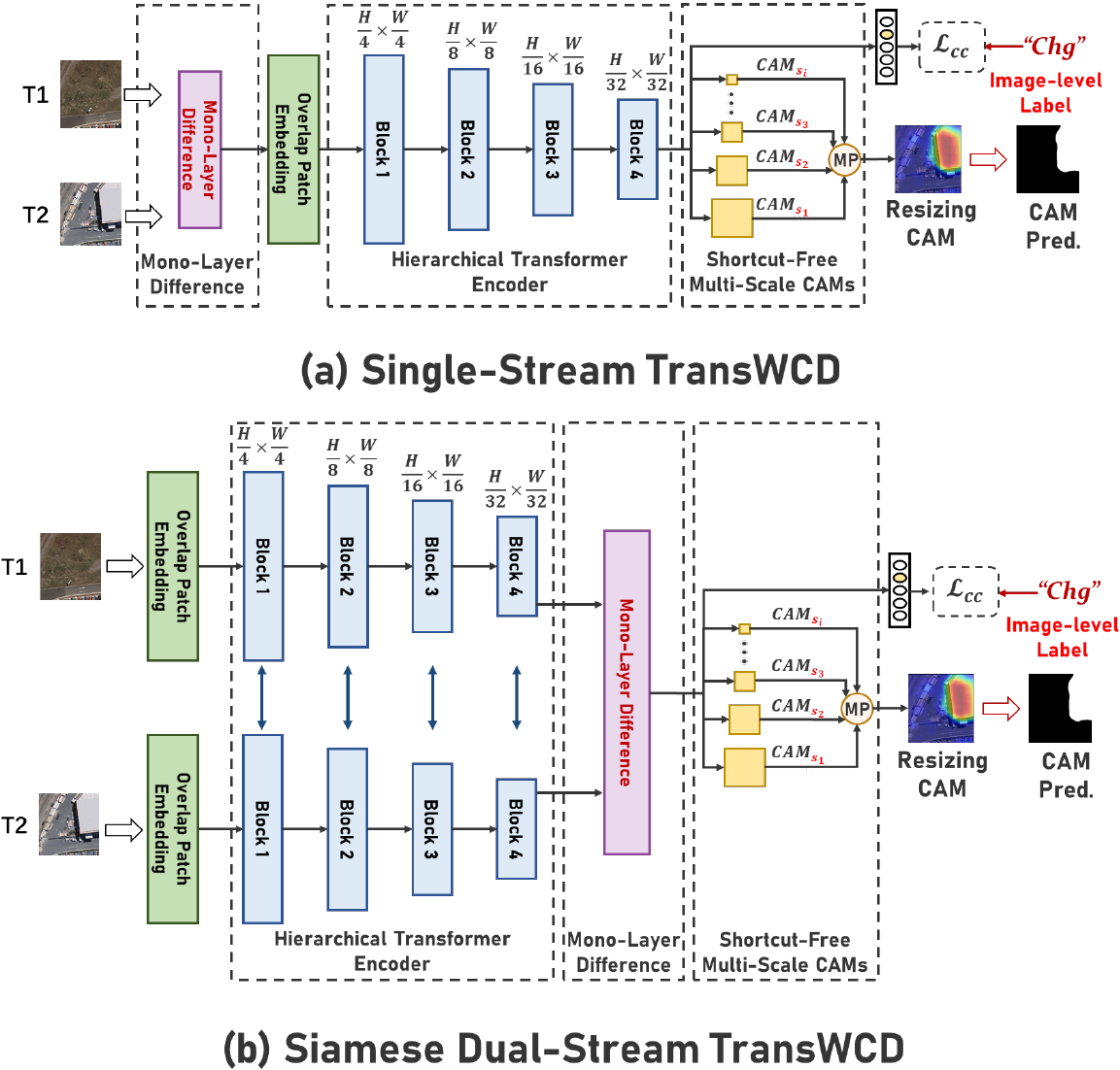}
	\caption{Pipeline of TransWCD. TransWCD is our basic model, including (a) single-stream and (b) Siamese dual-stream scheme. These schemes both comprise 1) hierarchical transformer encoder with image-level supervision, 2) mono-layer minimalist difference module, and 3) shortcut-free multi-scale CAM prediction module.}
	\label{method2}
\end{figure}
We strive to achieve an optimal balance between performance and complexity in constructing the basic model TransWCD. Our TransWCD consists of three main modules in Section \ref{sec：A}: 1) hierarchical transformer encoder with image-level supervision, 2) mono-layer minimalist difference module, and 3) shortcut-free multi-scale CAM prediction module. We explore two architectures, including single-stream and Siamese parameter-shared dual-stream. Both architectures share Module 1) and Module 3), but have different Module 2). The single-stream and dual-stream TransWCD pipelines are illustrated in Fig. \ref{method2}.

\begin{figure*}
	\centering
	\includegraphics[scale=0.55
]{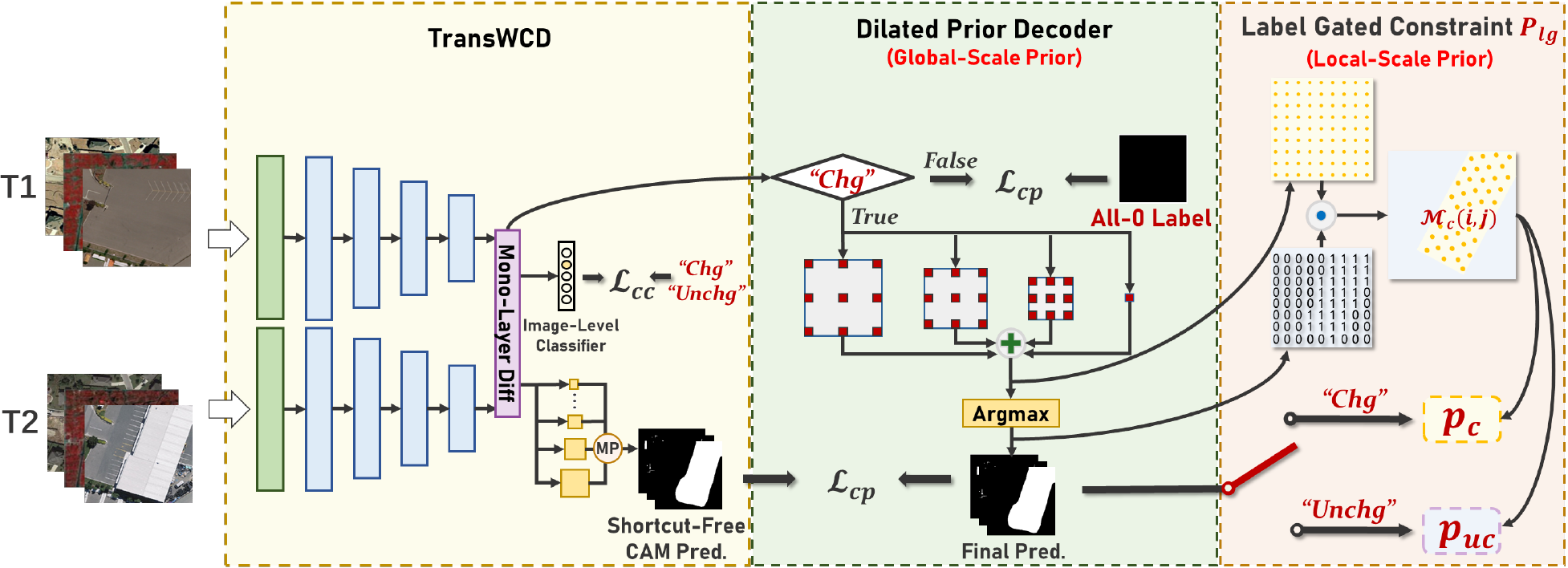}
	\caption{Framework of TransWCD-DL. TransWCD-DL consists of TransWCD, Dilated Prior (DP) decoder, and Label Gated (LG) constraint. DP decoder decodes the upstream feature map when the corresponding image-level labels are "Changed (Chg)." Otherwise, it utilizes an all-unchanged pixel-level label as the supervisory signal. LG constraint is a label-adaptive feature-level regularizer that relies on the changed features within the output feature map. These two modules incorporate the global-scale and local-scale priors to improve change missing and fabricating.}
	\label{method3}
\end{figure*}

Two additional components are integrated into TransWCD to cope with change missing and fabricating. This integration results in TransWCD-DL, as detailed in subsequent Section \ref{sec：B} and Section \ref{sec：C}. Section \ref{sec：B} introduces the dilated prior (DP) decoder incorporating the global-scale prior to address change fabricating, constructed of dilated convolution groups. Section \ref{sec：C} presents the label-gated (LG) constraint, which combines the local-scale prior to creating a feature-level penalty on change missing and fabricating. The LG constraint is calculated based on the relationship between image-level labels and the model's prediction of change status. Section \ref{sec: D} describes the end-to-end learning of TransWCD and TransWCD-DL. The overall framework of TransWCD-DL is depicted in Fig. \ref{method3}. 

\subsection{Weakly-Supervised Change Detection with Transformer}
\label{sec：A}
TransWCD incorporates two different architectures: the single-stream and Siamese dual-stream architectures. These architectures are determined based on the specific difference locations in pipeline. The single stream corresponds to the difference before encoder, and the dual stream is the encoder after difference. These two architectures contain different difference modules. To ensure a clear and concise description, we divide the hierarchical encoder with the mono-layer difference into two parts: the single stream and the dual stream. 

\subsubsection{Hierarchical Encoder with Mono-Layer Difference}
Our encoder-only TransWCD, which follows established CD pipelines \cite{guo2021deep,feng2023}, focuses on hierarchical feature extraction to capture multi-scale contextual information. Given a pre-change image $x_{t1}$ and a post-change image $x_{t2}$ with size $H\times W\times 3$.

{\bf{Single stream.}} As depicted in Fig. \ref{method2}(a), we initially feed the bi-temporal images into the difference module $D_s$:
\begin{equation}
\label{eq1}
x_d = Conv_{1 \times 1}(x_{t1}, x_{t2}).
\end{equation}
Note that no activation function is applied after the $Conv\ 1\times1$ operation. We only use $Conv\ 1\times1$ to reduce the dimension from 6 to 3, and refrain from unwanted nonlinearity that may affect the pixel relationship between paired nature images. For pre-trained models,  the absolute difference operation $|A-B|$ performs similarly to $Conv\ 1\times1$.

Next, the difference map $x_d$ with size $H\times W\times3$, $x_d$ is split into overlapping patches and entered into a four-stage hierarchical encoder with different resolutions $\left\{\frac{H}{4} \times \frac{W}{4}, \frac{H}{8} \times \frac{W}{8}, \frac{H}{16} \times \frac{W}{16}, \frac{H}{32} \times \frac{W}{32}\right\}$, resulting in channels $\{C_1,C_2,C_3,C_4\}$. At each stage, multi-head self-attention is computed among these patches to obtain multi-resolution difference representations $\{{Feat}_{d1},{Feat}_{d2},{Feat}_{d3},{Feat}_{d4}\}$. 

{\bf{Dual stream}}. In Fig. \ref{method2}(b), the bi-temporal images $x_{t1}$ and $x_{t2}$ are individually input into the hierarchical encoder, forming a Siamese structure with sharing parameters. Similar to the above description, the hierarchical encoder generates the multi-level features of $x_{t1}$ and $x_{t2}$, denoted as $\{{Feat}_{t11},{Feat}_{t12},{Feat}_{t13},{Feat}_{t14}\}$ and $\{{Feat}_{t21},{Feat}_{t22},{Feat}_{t23},{Feat}_{t24}\}$, respectively.
Next, we compute the difference result by differentiating the last-stage features ${Feat}_{t14}$ and ${Feat}_{t24}$:
\begin{equation}
\label{eq2}
Feat_{d4} = ReLU(Conv_{3 \times 3}(Feat_{t14}, Feat_{t24})).
\end{equation}
$Conv\ 3\times3$ is a standard convolution operation, followed by Rectified Linear Unit (ReLU) activation.

The design of the difference modules is plain and straightforward: $Conv\ 3\times3$ for the dual stream, and $Conv\ 1\times1$ without any activation functions for the single stream. $Conv\ 3\times3$ is utilized in the Siamese dual stream, while the single stream adopts $Conv\ 1\times1$ without activation functions. Despite their simplicity, these architectures exhibit exceptional performance compared to more complex designs.
\subsubsection{Image-Level Supervision}
Unlike FSCD, WSCD provides only image-level binary labels for training, instead of pixel-level ground truths, as depicted in Fig. \ref{introduction}. The label $y_{\text{cls}}^n$ indicates whether a pair is changed (1) or unchanged (0):
\begin{equation}
\label{eq3}
y_{\text{cls}}^n = \begin{cases}
1, & \text{\textit{if changed}} \\
0, & \text{\textit{if unchanged}}
\end{cases}.
\end{equation}

After obtaining the last-stage difference map $\text{Feat}_{d4}^n$ for the $n$-th paired images, we compute the change probability using a classifier, resulting in change probability vector $p_{\text{cls}}^n$. Then, we employ the binary cross-entropy with logits as change classification (CC) loss, optimizing the image-level change classification task and ensuring numerical stability during training:
\begin{equation}
\mathcal{L}_{\text{cc}}=\frac{1}{N}\sum_{n=1}^{N}\left[-y_{\text{cls}}^n\log{\sigma}(p_{\text{cls}}^n)-\left(1-y_{\text{cls}}^n\right)\log{\left(1-\sigma(p_{\text{cls}}^n)\right)}\right],
\label{eq_cc}
\end{equation}
where $N$ represents the batch quantity of bi-temporal images and $\sigma(\cdot)$ denotes the sigmoid activation function. 
\subsubsection{Shortcut-Free Multi-Scale CAM Prediction}
Class Activation Maps (CAMs) have proven effective for weakly-supervised dense prediction tasks \cite{yao2021, zhang2020causal}. By minor modifications to the last layer, CAMs can highlight the image regions most relevant to classification. Given their simplicity and interpretability, we leverage CAMs to generate initial pixel-level predictions of changed regions.

Specifically, after training the image-level classification layer, we employ the last-stage difference map $Feat_{d4}^n$, from the single- or dual-stream encoders, calculating the contribution of each pixel to the changed or unchanged class. Mathematically, this can be expressed as:
\begin{equation}
CAM^n=\sum_{i,j}^{h,w}W_{i,j}Feat_{d4}^n\left(i,j\right),
\end{equation}
where $CAM^n$ represents the class activation map for the n-th pair of images, and $(h,w)$ is the height and width of the feature map, respectively. $W_{i,j}$ denotes the weight associated with the pixel at position $(i,j)$ in the classification layer.

To incorporate multi-resolution presentation, we exploit multi-scale CAMs to predict change. These multi-scale CAMs are not generated from the multi-stage feature, but rather by scaling the last-stage feature $Feat_{d4}^n$. This last-stage approach offers computational efficiency and facilitates faster convergence. 

Precisely, resize the input images to different scales, denoted as $\{s_1, s_2, \ldots, s_S\}$, and compute multi-scale CAMs. Subsequently, upsample the CAMs to match the original input size $(H, W)$ and concatenate them, forming a list $\{CAM_{s_1}, CAM_{s_2}, \ldots, CAM_{s_S}\}$. Finally, sum and normalize the CAMs using adaptive max pooling $MP$, resulting in the final multi-scale CAMs:
\begin{equation}
\small
CAM_{\text{multi}}^n = \frac{\sum_{i=1}^{S} CAM_{s_i}^n}{MP\left(\sum_{i=1}^{S} CAM_{s_i}^n\right) + \epsilon},
\end{equation}
where $S$ represents the scale number, $CAM_{s_i}$ denotes the CAM for the $i$-th scale, and $\epsilon$ is a small positive constant added for numerical stability. A change score $\alpha \in \{0,1\}$ is introduced to discriminate the changed and unchanged regions. Then, we apply $Argmax$ to determine the change class with the highest probability at each pixel position, generating initial predictions $Pred_{\text{init}}^n \in \mathbb{R}^{H \times W}$:
\begin{equation}
\small
Pred_{\text{init}}^n = \begin{cases}
\text{\textit{Argmax}} \, [CAM_{\text{\textit{multi}}}^n], & \text{\textit{if} } \max \, [CAM_{\text{\textit{multi}}}^n] \geq \alpha \\
0, & \text{\textit{otherwise}}
\end{cases}.
\end{equation}

Following the principle of Occam's Razor, we minimize the redundant model complexity and prevent TransWCD from overfitting, resulting in a more competitive performance. The subsequent experimentation, ablation studies, and discussions demonstrate its superiority over more intricate structures. Due to its conciseness and effectiveness, TransWCD could serve as a strong baseline for future research in WSCD.

\subsection{Dilated Prior Decoder}
\label{sec：B}
\begin{figure}
	\centering
	\includegraphics[scale=0.25]{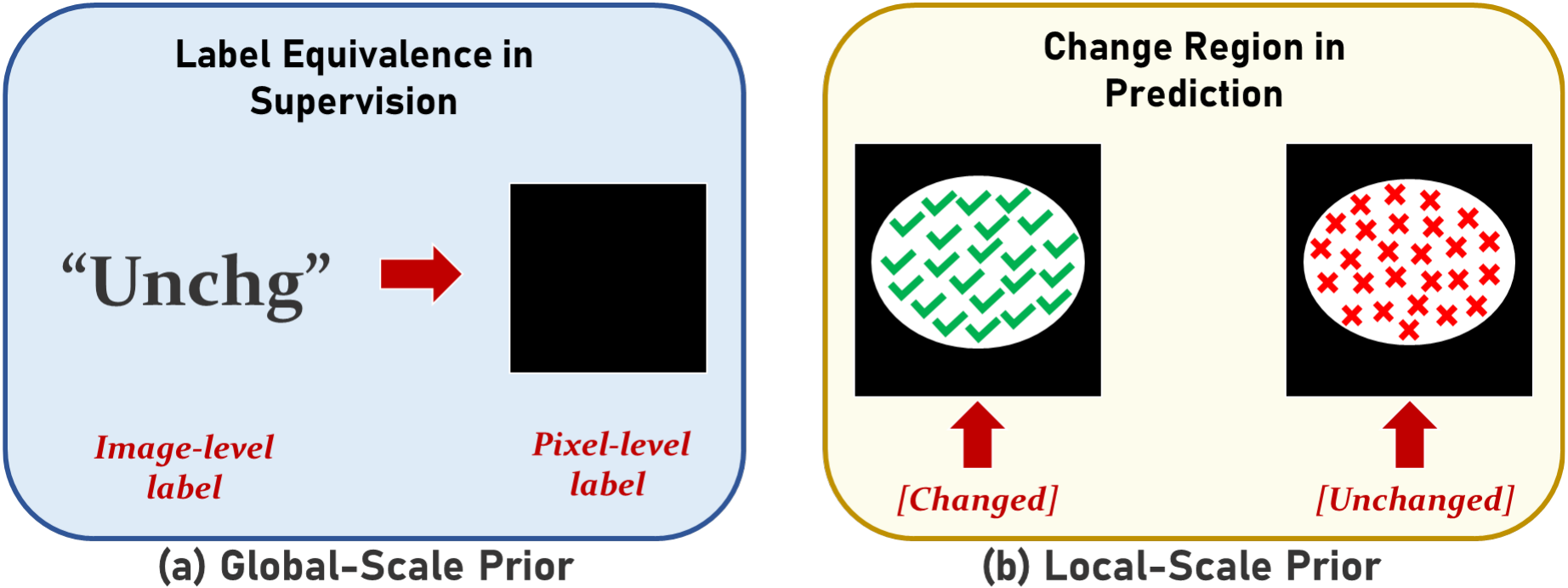}
	\caption{Global-scale prior and Local-scale prior. (a) The global-scale prior states that the "\textit{Unchanged (Unchg)}" image-level label is equivalent to all-zero pixel-level ground truth. (b) The local-scale prior states that if the image-level label is "\textit{Changed (Chg)}," there should be a presence of changed regions; otherwise, there must be an absence of changed regions.}	\label{method4}
\end{figure}

The image-level labels in WSCD inherently carry valuable global-scale prior, as illustrated in Fig. \ref{method4}(a):

\textit{When the image-level label is 0, indicating no change, it corresponds to all-unchanged pixel-level ground truth.}

We propose the Dilated Prior (DP) decoder integrating this global-scale prior, to address change fabricating, as depicted in Fig. \ref{method3}. DP decoder comprises dilated convolutions to capture a border contextual understanding \cite{chen2017rethinking}. Diverging from conventional designs, DP decoder modifies the decoding process, and no longer processes the output from the encoder $Feat_{d4}^n$ when the image-level label $y_{cls}^n = 0$. It exploits an all-unchanged pixel-level label $Y_0^n$, instead of the initial predictions $Pred_{init}^n$, as the supervising signal of DP decoder. This design ensures DP decoder incorporates the global-scale prior when no change is present, and guides the model to avert change fabricating:
\begin{equation}
Y_{pp}^n =
\begin{cases}
Y_0^n, & \text{\textit{if} } y_{cls}^n = 0 \\
Pred_{init}^n, & \text{\textit{if} } y_{cls}^n = 1 \\
\end{cases},
\end{equation}
where $Y_0 \in R^{H \times W}$ is a matrix of zeros. DP decoder consists of parallel three dilated convolutions $\{Dil_{r1}, Dil_{r2}, Dil_{r3}\}$ with different dilated rates $\{r1, r2, r3\}$ along with $Conv\ 1 \times 1$. These operations are applied in the last-stage difference map $Feat_{d4}^n$, analogous to the computation of CAMs. Formally, the output of DP decoder is defined as:
\begin{equation}
\begin{split}
Feat_{dp}^n = \text{Concat}\left[Dil_{r1}(Feat_{d4}^n), Dil_{r2}(Feat_{d4}^n), \right. \\
\left. Dil_{r3}(Feat_{d4}^n), Conv_{1\times1}(Feat_{d4}^n)\right].
\end{split}
\end{equation}

We supervise DP decoder using $Y_{pp}^n$ and compute a change prediction (CP) loss $\mathcal{L}_{cp}$ through pixel-level cross-entropy classification:
\
\begin{equation}
\small
\begin{split}
\mathcal{L}_{cp} = -\frac{1}{N}\sum_{n=1}^{N} \left[ \frac{1}{HW}\sum_{i=1,j=1}^{H,W} \left( Y_{pp}^n \log\left(\sigma\left(p_{dp}^n(i,j)\right)\right) \right. \right. \\
+ \left. \left. (1-Y_{pp}^n) \log\left(1-\sigma\left(p_{dp}^n(i,j)\right)\right) \right) \right],
\end{split}
\label{eq_cp}
\end{equation}

where $p_{dp}^n(i,j)$ is the corresponding pixel-level change probability of $Feat_{dp}^n$. Finally, we utilize the output of $p_{dp}^n(i,j)$ to derive the change predictions $Pred_{final}^n \in R^{H \times W}$:
\begin{equation}
Pred_{final}^n = \text{\textit{Argmax}}\left[p_{dp}^n(i,j))\right].
\end{equation}
By employing this selective decoding, DP decoder mitigates change fabricating and also reduces computational overhead. 
\subsection{Label Gated Constraint}
\label{sec：C}
The local-scale prior to the individual change predictions is represented as shown in Fig. \ref{method4}(b):

\textit{If the binary image-level label $y_{cls}^n = 1$, a changed region in ground truth definitely exists. Conversely, no region is changed if $y_{cls}^n = 0$.}

\begin{figure}
	\centering
	\includegraphics[scale=0.4]{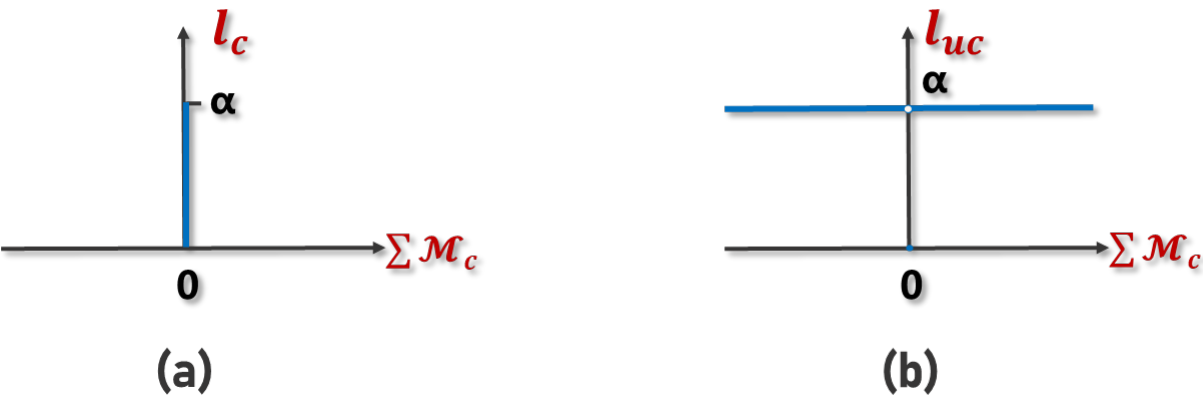}
	\caption{ Diagram illustrating the relationship between the changed constraint  $l_c$, the unchanged constraint $l_{uc}$, and the changed mask $M_c^n$ in LG constraint $L_{lg}$. The value of $l_c$ is $\alpha$ only when $\sum M_c^n[i,j] = 0$. In all other cases, $l_c$ is equal to 0. On the other hand, $l_{uc}$ takes a value of 0 when $\sum M_c^n[i,j] = 0$, and $\alpha$ otherwise. With this design, $l_c$ and $l_{uc}$ act as penalties in the case of change missing and fabricating, respectively.}
	\label{method5}
\end{figure}

This local-scale prior could assist in simultaneously ameliorating change missing and fabricating, and we propose a Label Gated (LG) constraint merging the observed knowledge. LG constraint serves as a penalty, triggered explicitly in worst-case scenario where mispredicting the change status. Formally, LG constraint imposes different penalty terms based on the image-level label $y_{cls}^n$:
\begin{equation}
L_{lg} = \frac{1}{N} \sum_{n=1}^{N} \left[y_{cls}^n l_c + (1-y_{cls}^n) l_{uc} \right].
\label{eq_lg}
\end{equation}
Here, $l_c$ and $l_{uc}$ are the regional feature-level constraints calculated from the output feature map. They are utilized to address the issue of change missing when the image-level label indicates changed $y_{cls}^n=1$, and mitigate the problem of change fabricating when the image-level label is unchanged $y_{cls}^n=0$, respectively.  As shown in Fig. \ref{method3}, we dynamically extract the changed region from the output feature map $Feat_{dp}^n$:
\begin{equation}
M_c^n=Pred_{final}^n \odot Feat_{dp}^n.
\end{equation}
Since $Pred_{final}^n$ contains binary pixel labels (0 or 1), element-wise multiplication $\odot$ allows us to obtain the changed region in $Feat_{dp}^n$. The changed mask $M_c^n$ identifies the changed pixels in $Feat_{dp}^n$:
\begin{equation}
\begin{aligned}
M_{c}^n(i,j) &= \begin{cases}
    Feat_{dp}^n(i,j), & \text{\textit{if} } Pred_{\text{\textit{final}}}^n[i,j] =1  \quad  \\
    0, & \text{\textit{otherwise}}
\end{cases}
\end{aligned},
\end{equation}
where $Pred_{init}^n[i,j] = 1$ represents a changed pixel at position $(i, j)$.  Next, we utilize $M_{c}^n(i,j)$ to sum the number of changed pixels in the output feature map. When $\sum M_c^n[i,j]=0$, it indicates that all elements in $M_c^n$ are 0, implying the absence of changed pixels in change prediction. If $y_{cls}^n=1$ in this case, indicating a change missing problem, we impose a penalty. Formally, we compute the changed constraint $l_c$ as follows:
\begin{equation}
l_c(M_c^n) = \alpha \cdot \delta\left[\sum M_c^n[i,j]=0\right],
\label{eq:pc}
\end{equation}
where $\alpha$ represents a weighting factor of the penalty term. As illustrated in Fig. \ref{method5}(a), $\delta[\cdot]$ is the unit impulse function that takes the value 1 only at the origin and 0 elsewhere. Conversely, if $\sum M_c^n[i,j] \neq 0$ in the case of $y_{cls}^n=1$, signifying a correct presence prediction of a changed region, $l_c(M_c^n)$ is set to 0 and has no impact on the model.

For $l_{uc}(M_c^n)$, it is utilized to penalize change fabricating where $y_{cls}^n = 0$, despite the presence of a changed region. We employ the complement of the unit impulse function $1-\delta(\cdot)$ to compute:
\begin{equation}
l_{uc}(M_c^n) = \alpha \cdot [1-\delta(\sum M_c^n[i,j]=0)].
\label{eq:puc}
\end{equation}
As decipted in Fig. \ref{method5}(b), when $\sum M_c^n[i,j] = 0$, it indicates that no change fabricating occurs, and there is a correct prediction of no change for all pixels. Thus, no penalty is imposed. When $\sum M_c^n[i,j] \neq 0$, it signifies that change fabricating exists, and in such cases, a penalty should be applied. To achieve this, we directly flip the calculation result using $1-\delta(\cdot)$, eliminating the need to compute $\sum M_c^n[i,j] \neq 0$.

The LG constraint $L_{lg}$ actively prompts the model to detect the presence of changed regions when $y_{cls}^n = 1$, and decisively penalizes false positive predictions when $y_{cls}^n = 0$. It is worth noting that LG constraint is not only employed in TransWCD-DL but also in enhancing TransWCD, where calculating regional features with $Pred_{init}^n$ instead of $Pred_{final}^n$.

To summarize, $L_{lg}$ enhances the model's capability to handle hard samples, mitigating the occurrences of change misssing and fabricating, and ensuring more reliable change prediction. Crucially, $L_{lg}$ could seamlessly integrate into the optimization process, without compromising the model's optimization on samples that accurately predict the change status.

\subsection{End-to-End Learning}
\label{sec: D}
We optimize both TransWCD and TransWCD-DL in an end-to-end manner. The foundational model, TransWCD, is trained with only the CC loss (Eq. \ref{eq_cc}):
\begin{equation}
\mathcal{L}_T = \mathcal{L}_{cc}.
\end{equation}
For TransWCD-DL, we merge LG constraint (Eq. \ref{eq_lg}) with the CC loss and CP loss (Eq. \ref{eq_cp}):
\begin{equation}
\mathcal{L}_{TD} = \mathcal{L}_{cc} + \varepsilon\mathcal{L}_{cp} + L_{lg}.
\label{eq_total_loss}
\end{equation}
Here, $\varepsilon$ is a scalar factor controlling the relative contribution of each component to the overall objective.

\section{Experiments}
\subsection{Experimental Settings}
\subsubsection{Dataset Description}
We conduct experiments on three widely available CD datasets: the learning, vision and remote sensing change detection (LEVIR-CD) \cite{chen2020spatial}, the Wuhan University building change detection (WHU-CD) \cite{ji2018fully}, and the profoundly supervised image fusion network change detection (DSIFN-CD) \cite{zhang2020deeply}. We adopt pre-processed datasets from previous works to ensure a fair comparison, rather than creating our own divisions in experiments. The images used in our experiments are all resized to 256×256 pixels. 

These three datasets are produced initially for FSCD without image-level labels. We offer a straightforward method for generating image-level labels, including in our open-source code, to enable the utilization of FSCD datasets in WSCD task.

{\bf{LEVIR-CD.}} This public large-scale building CD dataset consists of 637 pairs of high-resolution (0.5 m) remote sensing images of size 1024×1024. We adopt the default data split, where the original images are cropped into non-overlapping patches, including 4,538 changed samples and 5,654 unchanged samples \cite{han2023hanet}. The training/validation/test samples are 7120/1024/2048. 

{\bf{WHU-CD.} }This is an urban building CD dataset, containing one pair of high-resolution (0.3 m) aerial images of size 32,507×15,354, acquired in 2012 and 2016. We utilize the data from BGMix \cite{Huang2023}, which crops the image pairs with a 256×256 sliding window, and randomly select 90\% samples to train, and the left for testing. The cropped dataset totally contains 5,544 unchanged and 1,890 changed samples.

{\bf{DSIFN-CD.}} This dataset involves various land-cover objects, including roads, buildings, croplands, and water bodies. It comprises six pairs of high-resolution (2 m) satellite images from six cities. We exploit its default split, cropping to 14400/1360/192 samples for training/validation/test, consisting of 2,426 unchanged and 13,526 changed samples.

\subsubsection{Evaluation Protocol}
We employ the five most commonly used evaluation indicators \cite{31chen2021remote}: precision, recall, overall accuracy (OA), intersection over union (IoU), and F1 score. The F1 score, which combines precision and recall, is the primary metric for CD \cite{Zhang2022}.

\subsubsection{Implementation Details}
Unlike previous WSCD works that typically rely on ConvNets as encoders, we adopt a transformer encoder as the backbone. To optimize the model, we utilize the AdamW optimizer \cite{loshchilov2017}. The initial learning rate for the backbone is set to $5 \times 10^{-5}$ and decays using a polynomial scheduler. The learning rates for other modules, including the single-layer difference and DP decoder, are 10 times that of the backbone parameters. 

We train TransWCD for 30,000 iterations, employing a warm-up strategy. LG constraint is applied since the initial training phase, and DP decoder is integrated into the backpropagation process after 2,000 iterations. The batch size is 8, and the input size of bi-temporal images is $256 \times 256$. We employ random rescaling, horizontal flipping, and cropping to augment the data. The loss contribution factor $\varepsilon$ in Eq. \ref{eq_total_loss} is set to 0.1. The change score for the prediction from CAMs is empirically set to $\alpha = 0.45$. The scale group for multi-scale CAMs is $\{0.5, 1.0, 1,5,2.0\}$.

\subsection{Ablation Study of TransWCD}
In this section, we conduct extensive exploration and ablative experiments to determine the optimal structure for TransWCD. TransWCD refers to the single-stream scheme by default in the remaining experimental sections unless otherwise specified. 
\subsubsection{Analysis of TransWCD Backbone}
We develop various transformer encoders for our TransWCD with the goal of "less complexity but more competitive performance." As shown in Table \ref{encoder}, these encoder backbones are ViT-T and ViT-S \cite{dos2020image}; Swin-T \cite{liu2021swin}; PVT-T \cite{2021pyramid}; and MiT-B0, MiT-B1, and MiT-B2 \cite{segformer}. To comprehensively compare performance and efficiency, we evaluate using F1 score (\%), parameters (MB), and complexity (GFLOPs).

%\captionsetup[table]{labelformat=simple, labelsep=newline, textfont=sc,justification=centering}
\begin{table}
\setlength{\tabcolsep}{5pt}
\renewcommand{\arraystretch}{1.5}
\centering
\caption{Comparison of different encoders for TransWCD on WHU-CD dataset. Parameters (M), Flops (G), F1 score (\%), OA (\%), and IoU (\%) are reported.}
\begin{tabular}{cccccc} 
\hline\hline
\begin{tabular}[c]{@{}c@{}}\textbf{Transformer }\\\textbf{Encoder}\end{tabular} & \textbf{Params (M) ↓} & \textbf{Flops (G) ↓} & \textbf{F1 ↑} & \textbf{OA ↑} & \textbf{IoU ↑}  \\ 
\hline
\textbf{ViT-T}                                                                  & 5.72              & 1.58             & 63.14       & 97.25       & 45.75         \\
\textbf{ViT-S}                                                                  & 22.13             & 5.97             & 66.98       & 95.78       & 50.04         \\
\textbf{Swin-T}                                                                 & 28.34             & 8.15             & 67.46       & 96.61       & 51.55         \\
\textbf{PVT-T}                                                                  & 13.22             & 3.74             & 65.97       & 95.25       & 47.40         \\
\textbf{MiT-B0}                                                                 & 3.32              & 0.93             & 62.37       & 94.60       & 44.55         \\
\textbf{MiT-B1}                                                                 & 13.15             & 3.60             & 67.81       & 97.01       & 50.80         \\
\textbf{MiT-B2}                                                                 & 24.20             & 6.22             & 67.39       & 97.16       & 51.30         \\
\hline\hline
\end{tabular}
\label{encoder}
\end{table}

The ViT series share the same settings, including a patch size of $16 \times 16$ with overlapping. Swin-T, PVT-T, and the MiT series have four-stage hierarchical structures. It is observed that MiT-B2 and Swin-T have similar model sizes to ViT-S, but outperform ViT-S with F1 scores of +0.41\% and +0.48\%, respectively. This result demonstrates that the hierarchical transformer structures contribute to improvements in WSCD to some extent. MiT-B1 exhibits the best tradeoff between parameters and performance among all these backbones.  MiT-B1 even outperforms other MiTs in both performance and complexity, showing a significant advantage. For instance, MiT-B1 surpasses MiT-B2 in performance and parameter efficiency, with a +0.42\% F1 score and -11.05MB parameters. This discrepancy could be attributed to the excessive parameters of MiT-B2, which may cause TransWCD to learn unnecessary or noisy information, ultimately reducing its generalization ability \cite{zhang2021understanding}.

\subsubsection{Analysis of TransWCD Structure}
We explore several schemes regarding the location and structure of difference for TransWCD. The difference module is positioned before or after the encoder, forming a single-stream or Siamese dual-stream structure. As shown in Table \ref{difference}, single-stream differences include $Diff(A, B)$ , $|A-B|$,  $Conv\ 1\times1$, and $Conv\ 1\times1\ w/o\ ReLu$; dual-stream differences contain  $Conv\ 1\times1$, $Conv\ 3\times3$, and $Diff(A, B)$.

Overall, the dual-stream structure outperforms the single-stream structure. The best-performing double-stream configuration achieves a +0.92\% F1 score and a +1.06\% IoU compared to its single-stream counterpart.  Specifically, within the dual-stream setup, a single layer of $Conv\ 3\times3$ performs best among the options considered. As for the single-stream setup, there is a comparable performance between $|A-B|$ and $Conv\ 1\times1\ w/o\ ReLu$. Interestingly, removing the non-linear operation leads to a +5.40\% F1 score and +5.94\% IoU. We speculate that the non-linear operation may alter the pixel relationships between images after the difference operation, or this phenomenon is related to using pre-trained parameters.
\begin{table}
\setlength{\tabcolsep}{3.8pt}
\renewcommand{\arraystretch}{1.6}
\centering
\caption{Ablation study of different difference modules for TransWCD on WHU-CD dataset. $Diff(A, B)$ represents two layers of $Conv\ 3\times3$. F1 score (\%), precision (\%), recall (\%), OA (\%), and IoU (\%) are reported.}
\begin{tabular}{ccccccc} 
\hline\hline
\multicolumn{2}{c}{\textbf{Difference Structure}}                                  & \textbf{F1 ↑} & \textbf{Pre. ↑} & \textbf{Rec. ↑} & \textbf{OA ↑} & \textbf{IoU ↑}  \\ 
\hline
\multirow{4}{*}
{\begin{tabular}[c]{@{}c@{}}\textbf{Single}\\\textbf{Stream}\end{tabular}} & Diff (A, B)                & 52.84                                           & 42.76                                             & 69.14         & 95.10                                           & 35.91                                             \\
                                                                                          & \textbar{}A-B\textbar{}    & 67.58                                           & 64.25                                             & 71.28         & 97.29                                           & 51.04                                             \\
                                                                                          & Conv 1×1\textbf{~}         & 62.41                                           & 53.29                                             & 75.28         & 96.40                                           & 45.36                                             \\
                                                                                          & Conv 1×1 \textit{\small w/o ReLu} & 67.81                                           & 61.67                                             & 75.31         & 97.16                                           & 51.30                                             \\                          \hline                        
\multirow{3}{*}
{\begin{tabular}[c]{@{}c@{}}\textbf{Dual}\\\textbf{Stream}\end{tabular}}   & Conv 1×1~                  & 64.63                                           & 71.15                                             & 59.20         & 96.81                                           & 47.74                                             \\
                                                                                          & Conv 3×3~                  & 68.73                                           & 75.34                                             & 63.19         & 97.17                                           & 52.36                                             \\
                                                                                          & Diff (A, B)                & 67.70                                           & 72.52                                             & 63.49         & 97.02                                           & 51.17                                             \\
\hline\hline
\end{tabular}
\label{difference}
\end{table}
\begin{table}
\setlength{\tabcolsep}{8pt}
\renewcommand{\arraystretch}{1.6}
\centering
\caption{Effectiveness of DP Decoder and LG Constraint on WHU-CD. F1 score (\%) is reported.}
\begin{tabular}{ccccc} 
\hline\hline
\multicolumn{2}{c}{\textbf{Method}}                                                                                       & \begin{tabular}[c]{@{}c@{}}\textbf{DP }\\\textbf{Decoder}\end{tabular} & \begin{tabular}[c]{@{}c@{}}\textbf{LG } \\\textbf{Constraint}\end{tabular} & \textbf{ F1 ↑}                               \\ 
\hline
\multirow{4}{*}{\begin{tabular}[c]{@{}c@{}}\textbf{\textbf{Single}}\\\textbf{\textbf{Stream}}\end{tabular}} & TransWCD    & ~                                                                      & ~                                                                          & 67.81                                      \\
                                                                                                            & TransWCD-D  & \textbf{\checkmark}                                                              & ~                                                                          & 69.88                                      \\
                                                                                                            & TransWCD-L  & ~                                                                      & \textbf{\checkmark}                                                                  & 68.45                                      \\
                                                                                                            & TransWCD-DL & \textbf{\checkmark}                                                              & \textbf{\checkmark}                                                                  & 70.67                                      \\
                                                    \hline
\multirow{4}{*}{\begin{tabular}[c]{@{}c@{}}\textbf{\textbf{Dual}}\\\textbf{\textbf{Stream}}\end{tabular}}   & TransWCD    & \textbf{~}                                                             & \textbf{~}                                                                 & {68.73}  \\
                                                                                                            & TransWCD-D  & \textbf{\checkmark}                                                              & \textbf{~}                                                                 & 71.02                                      \\
                                                                                                            & TransWCD-L  & ~                                                                      & \textbf{\checkmark}                                                                  & 69.54                                      \\
                                                                                                            & TransWCD-DL & \textbf{\checkmark}                                                              & \textbf{\checkmark}                                                                  & 71.95                                      \\
\hline\hline
\end{tabular}
\label{DP_LG}
\end{table}

\subsection{Ablation Study of DP Decoder and LG Constraint}
\subsubsection{Analysis of Component Validity}
\begin{figure*}[!t]
	\centering
	\includegraphics[scale=0.54]{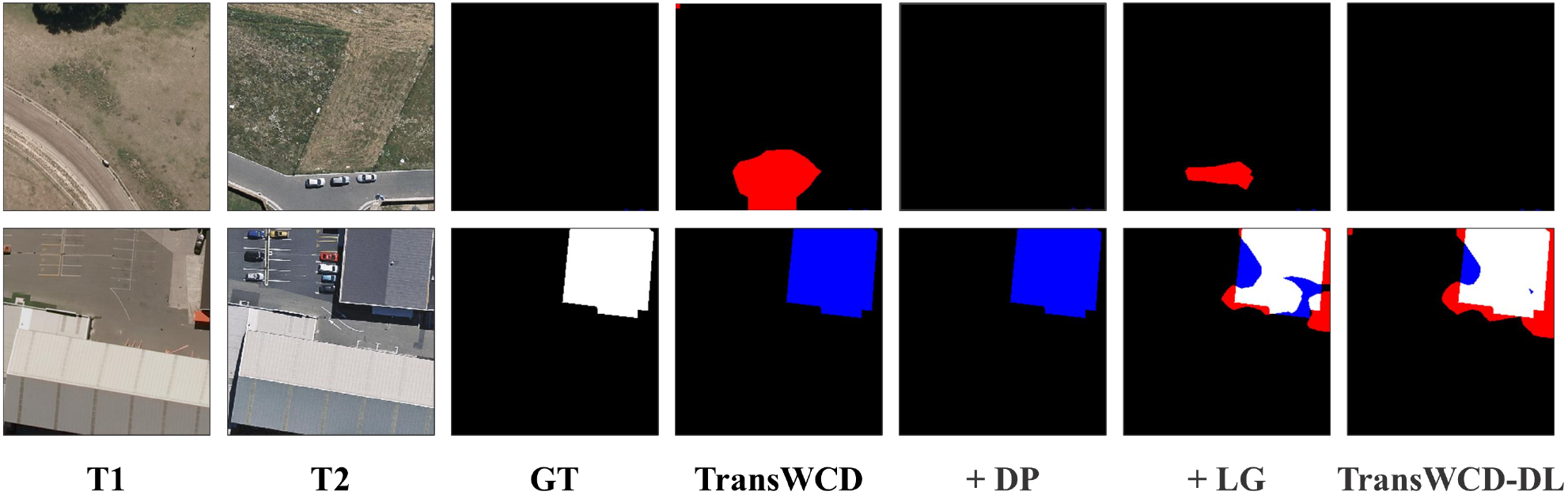}
 \caption{Improvements by DP decoder and LG constraint. \textcolor{red}{False positives} and \textcolor{blue}{false negatives} are highlighted in red and blue, respectively. It is evident from the 1st line that DP decoder eliminates the sample's change fabricating, and LG constraint reduces the area of change fabricating. Furthermore, as indicated in the 2nd line, LG constraint addresses the issue of change missing.} \label{experiment1}
\end{figure*}
\begin{figure}
	\centering
	\includegraphics[scale=0.49]{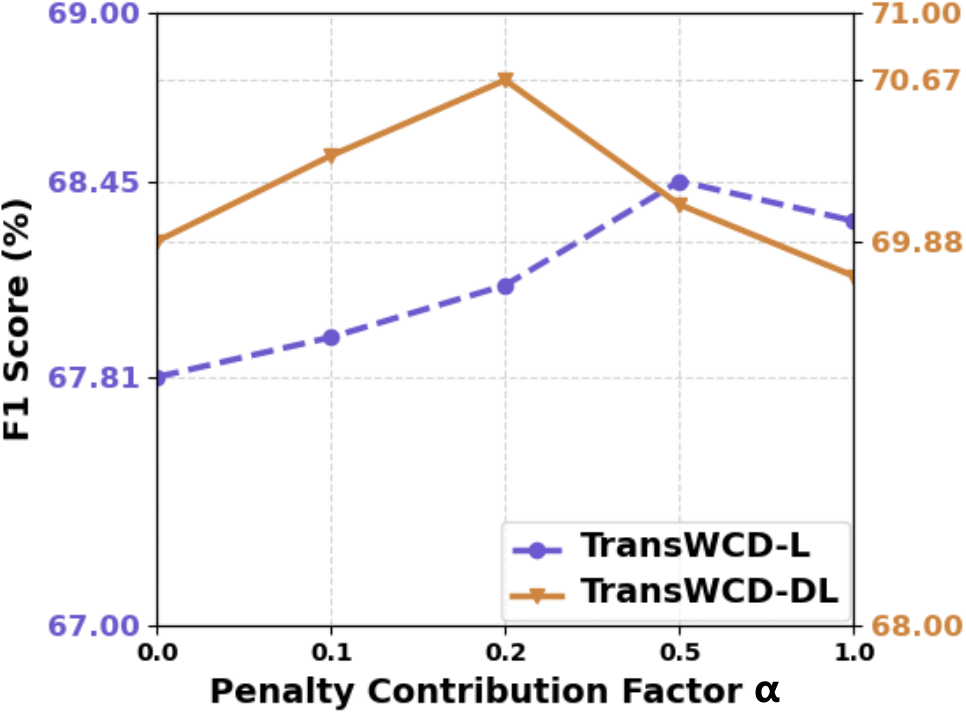}
	\caption{Sensitivity analysis results of penalty contribution factor $\alpha$ for LG constraint $L_{lg}$ on WHU-CD dataset. F1 score (\%) is reported.}
	\label{experiment2}
\end{figure}The effectiveness of the proposed DP decoder and LG constraint is demonstrated in Table \ref{DP_LG}. DP decoder shows a significant improvement in F1 score, with a +2.07\% for the single-stream architecture and a +2.29\% for the dual-stream architecture, compared to the basic model TransWCD on WHU-CD dataset. Applying LG constraint directly to TransWCD leads to a +0.64\% F1 score for the single-stream models and a +0.81\% F1 score for the dual-stream models. When combined with DP decoder, LG constraint further boosts performance, achieving a +0.79\% F1 score for single-stream models and a +0.93\% F1 score for dual-stream models. The integration of DP decoder on top of LG constraint yields a +2.22\% and +2.71\% improvement in F1 score for the single-stream and dual-stream models, respectively. These experimental findings solidify the positive impact of DP decoder and LG constraint in improving change missing and fabricating, and boosting the reliability of prediction.

We further visualize the prediction improvements brought by DP decoder and LG constraint, as depicted in Fig. \ref{experiment1}. In the 1st row, it is observed that for the unchanged bi-temporal images, TransWCD erroneously predicts the presence of change regions (change fabricating). However, by incorporating DP decoder, the issue of false positives is successfully addressed, providing direct evidence for the effectiveness of DP decoder to converting the unchanged image-level labels into all-unchanged pixel-level predictions. Additionally, the integration of LG constraint alleviates the issue of change fabricating, reducing the occurrence of false positive pixels, and highlighting the capability to refine the prediction of change regions under unchanged labels.

Moving to the 2nd row, TransWCD fails to detect the corresponding change regions for the changed bi-temporal images (change missing). By incorporating LG constraint, however, the change predictions are significantly improved. This improvement indicates the capacity of LG constraint to rectify undetected change regions under the changed labels, i.e., ameliorating the misprediction of change status.

\subsubsection{Analysis of DP Decoder Structure}
\begin{table}
\setlength{\tabcolsep}{8pt}
\renewcommand{\arraystretch}{1.7}
\centering
\caption{Comparison of different dilated rate groups for DP decoder on WHU-CD. F1 score (\%), Precision (\%), Recall (\%), OA (\%), and IoU (\%) are reported.}
\begin{tabular}{cccccc} 
\hline\hline
\textbf{\textbf{Dilated Rate}} & \textbf{F1. ↑} & \multicolumn{1}{l}{\textbf{\textbf{Pre. ↑}}} & \textbf{Rec. ↑} & \textbf{\textbf{OA ↑}}~ & \textbf{IoU ↑}~  \\ 
\hline
\textbf{TransWCD}              & 67.81        & 61.67                                      & 75.31         & 97.16                 & 51.30          \\
\textbf{[0, 2, 4, 8]}\textit{} & 61.60        & 53.92                                      & 71.85         & 96.45                 & 44.51         \\
\textbf{[1, 2, 3, 4]}          & 68.54        & 64.31                                      & 73.37         & 97.33                 & 52.14          \\
\textbf{[0, 1, 2, 3]}          & 69.88        & 68.61                                      & 71.19         & 97.56                 & 53.70          \\
\hline\hline
\end{tabular}
\label{DP}
\end{table}
We conduct experiments on different dilated rate groups for DP decoder to determine the optimal design, as presented in Table \ref{DP}. DP decoder exploits the last-stage feature map $Feat_{d4}^n$ to obtain multi-scale features through dilation. The notation $[0,1,2,3]$ represents a configuration where DP decoder consists of one $Conv\ 1\times1$ layer and three $Conv\ 3\times3$ layers, with dilated rates set as $[1,2,3]$, where 0 denotes $Conv\ 1\times1$. Similarly,  $[0,2,4,8]$ denotes the group of a $Conv\ 1\times1$ and  three $Conv\ 3\times3$ with dilated rates $[2,4,8]$, among others.

DP decoder with the configuration of $[0,1,2,3]$ demonstrates the best performance, surpassing the $[1,2,3,4]$ and $[0,2,4,8]$ configurations by +1.34\% and +8.28\% in terms of F1 score, and +1.56\% and +9.19\% in terms of IoU. It is evident that lower dilation rates yield better results, whereas excessively high dilation rates result in performance degradation. DP decoder with $[0,2,4,8]$ even exhibits a -6.21\% in F1 score and a -6.79\% IoU, compared to the original TransWCD model. 
These observations can be attributed to the relatively small-scale characteristics of remote-sensing objects. More significant dilation rates, which imply a larger receptive field, may lead to a loss of fine-grained details and spatial resolution for remote-sensing objects. Moreover, larger dilation rates may introduce more noise and irrelevant features.

\subsubsection{Sensitivity Analysis of $\boldsymbol{\alpha}$}
\begin{table*}
\setlength{\tabcolsep}{10pt}
\renewcommand{\arraystretch}{1.5}
\centering
\caption{Quantitative comparison with weakly-supervised change detection methods. The performance metrics exceeding the WSCD state-of-the-art are highlighted in red. F1 score (\%), Precision (\%), Recall (\%), OA (\%), IoU (\%) and Venue are reported.}
\begin{tabular}{ccccccccc} 
\hline\hline
\multicolumn{2}{c}{\multirow{2}{*}{\textbf{Method}}}                                                                 & \multirow{2}{*}{\textbf{Backbone}} & \multicolumn{5}{c}{\textbf{WHU-CD}}                                                                                                                                     & \multirow{2}{*}{\textbf{Venue}}  \\ 
\cline{4-8}
\multicolumn{2}{c}{}                                                                                                 &                                    & \textbf{F1 ↑}                     & \textbf{Pre. ↑}                   & \textbf{Rec. ↑}                   & \textbf{OA ↑}                     & \textbf{IoU ↑}                    &                                  \\ 
\hline
\multirow{7}{*}{\begin{tabular}[c]{@{}c@{}}\textbf{Weakly}\\\textbf{Supervised}\end{tabular}} & WCDNet \cite{ander2020}             & \multirow{5}{*}{ConvNet}           & 39.30                           & -                               & -                               & 82.30                           & 22.10                           & ACCV 2020                        \\
                                                                                              & Kalita \textit{et al.} \cite{kal2021}       &                                    & 32.45                           & 23.87                           & 50.67                           & 91.63                           & 19.37                           & CAIP 2021                        \\
                                                                                              & CAM \cite{wu2023}              &                                    & 54.60                           & 49.40                           & 61.02                           & 79.29                           & 37.55                           & -                      \\
                                                                                              & FCD-GAN \cite{wu2023}          &                                    & 56.45                           & 48.93                           & 66.78                           & 78.98                           & 39.32                           & TPAMI 2023                       \\
                                                                                              & BGMix \cite{Huang2023}              &                                    & 62.40                           & -                               & -                               & 84.40                           & 42.70                           & AAAI 2023                        \\ 
\cline{2-9}
                                                                                              & \textbf{TransWCD}    & \multirow{2}{*}{Transformer}       & \textbf{68.73}                  & \textbf{\textcolor{red}{75.34}} & \textbf{63.19 }                 & \textbf{97.17}                  & \textbf{52.36}                  & \textbf{-}                       \\
                                                                                              & \textbf{TransWCD-DL} &                                    & \textbf{\textcolor{red}{71.95}} & \textbf{64.46}                  & \textbf{\textcolor{red}{81.42}} & \textbf{\textcolor{red}{98.34}} & \textbf{\textcolor{red}{56.19}} & \textbf{-}                       \\
\hline\hline
\end{tabular}
\label{WSSS}
\end{table*}
We investigate the impact of varying the penalty contribution factor $\alpha$ of LG constraint $L_{lg}$ on TransWCD-L and TransWCD-DL, as depicted in Fig. \ref{experiment2}. For TransWCD-L, it is observed that an appropriate $\alpha$ value within the range of $[0, 0.5]$ consistently improves the performance. When $\alpha=0.5$, TransWCD-L achieves its optimal performance with an F1 score of 68.45\%. Similarly, TransWCD-DL demonstrates its best performance with an F1 score of 70.67\%, when utilizing a smaller $\alpha$ value of 0.2. 

However, it is worth noting that larger $\alpha$ values result in performance degradation, and in extreme cases, when $\alpha =1.0$, it even leads to a negative impact. This phenomenon arises due to the strong constraint imposed by larger $\alpha$ values, leading to the suppression of CC loss $\mathcal{L}_{cc}$ and CP loss $\mathcal{L}_{cp}$. Consequently, the model's capability to accurately perform image-level and pixel-level classifications is constrained.
\begin{figure*}[!t]
	\centering
	\includegraphics[scale=0.6]{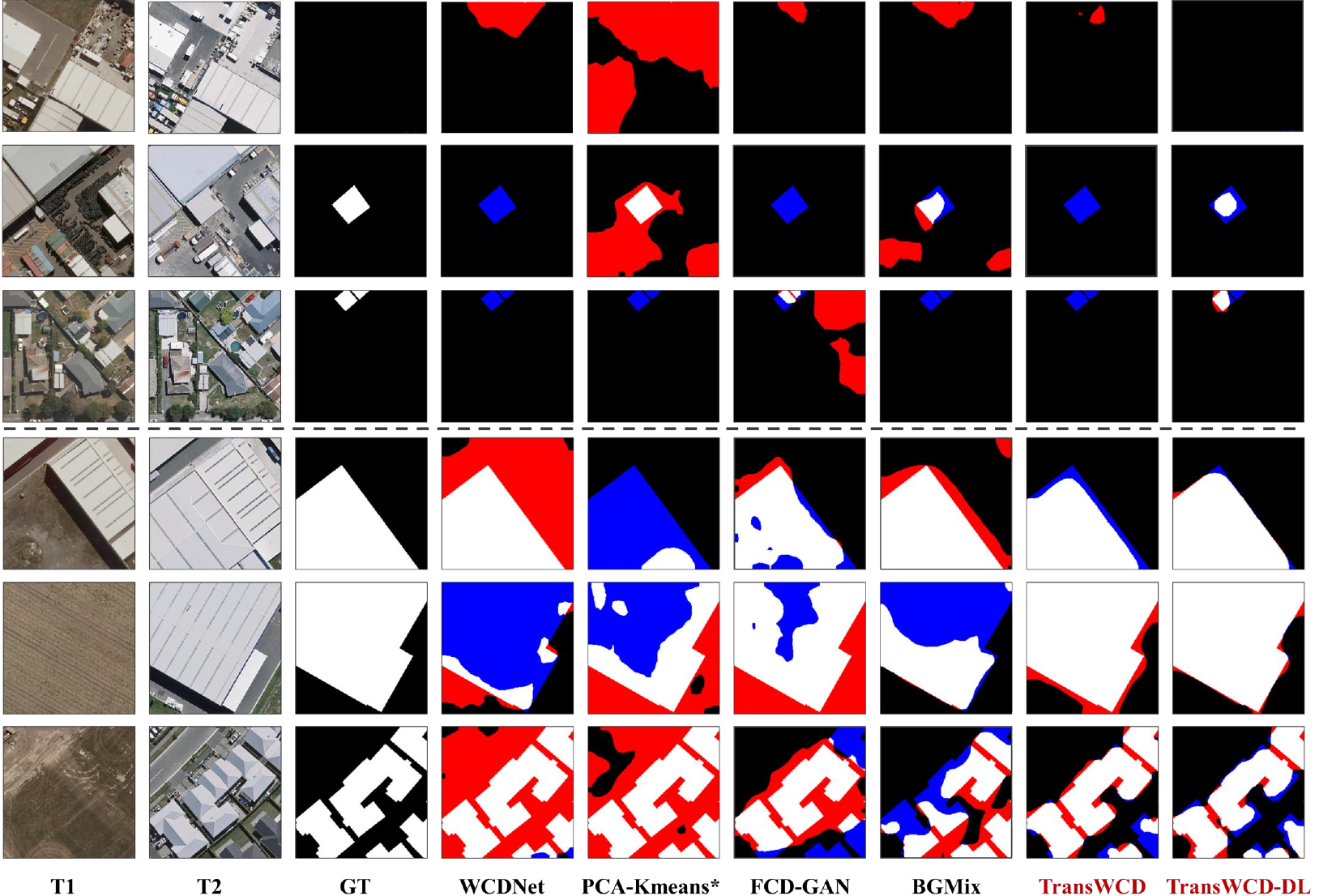}
	\caption{Qualitative comparison of our TransWCD and TransWCD-DL with other weakly-supervised change detection methods. \textcolor{red}{False positives} and \textcolor{blue}{false negatives} are highlighted in red and blue, respectively. PCA-Kmeans* is from \cite{kal2021}. The 1st row clearly shows the enhancement of TransWCD-DL in addressing the issue of change fabricating. The 2nd and 3rd rows demonstrate the correction of change missing by TransWCD-DL. The subsequent three rows validate the performance improvement of TransWCD and TransWCD-DL in predicting general samples, compared to existing WSCD methods.}
	\label{experiment3}
\end{figure*}

\subsection{Comparison to State-of-the-Art Methods}

 %\captionsetup[table]{labelformat=simple, labelsep=newline, textfont=sc,justification=centering}
\begin{table*}
\setlength{\tabcolsep}{2.7pt}
\renewcommand{\arraystretch}{1.5}
\centering
\caption{Comparison with state-of-the-art fully-supervised change detection methods. The performance metrics that surpass those of one FSCD method are highlighted in red. F1 score (\%), Precision (\%), Recall (\%), OA (\%), and IoU (\%) are reported.}
\begin{tabular}{cccccccccccccccccccc} 
\hline\hline
\multicolumn{2}{c}{\multirow{2}{*}{\textbf{Method}}}                                                                                   & \multicolumn{5}{c}{\textbf{LEVIR-CD}}                                              & \multirow{2}{*}{} & \multicolumn{5}{c}{\textbf{WHU-CD}}                                                                                                                                                                                                                                                 & \multirow{2}{*}{} & \multicolumn{5}{c}{\textbf{DSIFN-CD}}                                                                                                  \\ 
\cline{3-7}\cline{9-13}\cline{15-19}
\multicolumn{2}{c}{}                                                                                                                   & \textbf{F1 ↑}    & \textbf{Pre. ↑}  & \textbf{Rec. ↑}  & \textbf{OA ↑}    & \textbf{IoU ↑}   &                   & \textbf{F1 ↑}                                                         & \textbf{Pre. ↑}                   & \textbf{Rec. ↑}                   & \textbf{OA ↑}                                                         & \textbf{IoU ↑}                                                        &                   & \textbf{F1 ↑}    & \textbf{Pre. ↑}                   & \textbf{Rec. ↑}                   & \textbf{OA ↑}                     & \textbf{IoU ↑}                                  \\ 
\hline
\multirow{7}{*}{\begin{tabular}[c]{@{}c@{}}\textbf{Fully}\\\textbf{Supervised}\end{tabular}}                    & FC-EF \cite{47daudt2018fully}           & 83.40          & 86.91          & 80.17          & 98.39          & 71.53          & ~                 & 69.37                                                               & 71.63                           & 67.25                           & 97.61                                                               & 53.11                                                               & ~                 & 61.09          & 72.61                           & 52.73                           & 88.59                           & 43.98                                      \\
                                                                                                                & FC-Siam-Di \cite{31chen2021remote}        & 86.31          & 89.53          & 83.31          & 98.67          & 75.92          & ~                 & 58.81                                                               & 47.33                           & 77.66                           & 95.63                                                               & 41.66                                                               & ~                 & 62.54          & 59.67                           & 65.71                           & 86.63                           & 45.50                                      \\
                                                                                                                & FC-Siam-Conc \cite{31chen2021remote}      & 83.69          & 91.99          & 76.77          & 98.49          & 71.96          & ~                 & 66.63                                                               & 60.88                           & 73.58                           & 97.04                                                               & 49.95                                                               & ~                 & 59.71          & 66.45                           & 54.21                           & 87.57                           & 42.56                                      \\
                                                                                                                & STANet \cite{34bandara2020spatial}          & 87.26          & 83.81          & 91.00          & 98.66          & 77.40          & ~                 & 82.32                                                               & 79.37                           & 85.50                           & 98.52                                                               & 69.95                                                               & ~                 & 64.56          & 67.71                           & 61.68                           & 88.49                           & 47.66                                    \\
                                                                                                                & IFNet \cite{35zhang2020deeply}           & 88.13          & 94.02          & 82.93          & 98.87          & 78.77          & ~                 & 83.40                                                               & 96.91                           & 73.19                           & 98.83                                                               & 71.52                                                               & ~                 & 60.10          & 67.86                           & 53.94                           & 87.83                           & 42.96                                    \\
                                                                                                                & DTCDSCN \cite{40liu2021building}         & 87.67          & 88.53          & 86.83          & 98.77          & 78.05          &                   & 71.95                                                               & 63.92                           & 82.30                           & 97.42                                                               & 56.19                                                               &                   & 62.72          & 53.87                           & 77.99                           & 84.91                           & 46.76                                    \\
                                                                                                                & SNUNet \cite{37fang2022snunet}         & 88.16          & 89.18          & 87.17          & 98.82          & 78.83          & ~                 & 83.50                                                               & 85.60                           & 81.49                           & 98.71                                                               & 71.67                                                               & ~                 & 66.18          & 60.60                           & 72.89                           & 87.34                           & 49.45                                     \\ 
\hline
\multirow{2}{*}{\begin{tabular}[c]{@{}c@{}}\textbf{\textbf{Weakly}}\\\textbf{\textbf{Supervised}}\end{tabular}} & \textbf{TransWCD}    & \textbf{60.08} & \textbf{55.48} & \textbf{65.51} & \textbf{95.56} & \textbf{52.94} &                   & \textbf{\textcolor{red}{68.73}}                                     & \textbf{\textcolor{red}{75.34}} & \textbf{65.19}                  & \textbf{\textcolor{red}{97.17}}                                     & \textbf{\textcolor{red}{52.36}}                                     &                   & \textbf{53.41} & \textbf{\textcolor{red}{80.19}} & \textbf{\textcolor{red}{71.39}} & \textbf{61.05}                  & \textbf{36.44}                      \\
                                                                                                                & \textbf{TransWCD-DL} & \textbf{63.29} & \textbf{60.85} & \textbf{65.93} & \textbf{96.10} & \textbf{56.30} &                   & \textbf{\textcolor{red}{71.95}\textcolor[rgb]{0.212,0.208,0.208}{}} & \textbf{\textcolor{red}{64.46}} & \textbf{\textcolor{red}{81.42}} & \textbf{\textcolor{red}{98.34}\textcolor[rgb]{0.212,0.208,0.208}{}} & \textbf{\textcolor{red}{56.19}\textcolor[rgb]{0.212,0.208,0.208}{}} &                   & \textbf{56.22} & \textbf{47.95}                  & \textbf{67.92}                  & \textbf{\textcolor{red}{84.94}} & \textbf{39.10}                     \\
\hline\hline
\end{tabular}
\label{FSSS}
\end{table*}

\subsubsection{Weakly-Supervised Change Detection}
We compare the performance of TransWCD and our whole algorithm TransWCD-DL with the existing WSCD works, including FCD-GAN \cite{wu2023}, CAM \cite{wu2023}, BGMix \cite{Huang2023}, WCDNet \cite{ander2020}, and Kalita \textit{et al.} \cite{kal2021}. CAM \cite{wu2023} is a basic CAM method, and the performance of \cite{kal2021} is our replication using MiT-B1 encoder.

The quantitative results on WHU-CD datasets are shown in Table \ref{WSSS}. Even our TransWCD remarkably outperforms other WSCD methods by at least a +6.33\% F1 score, without bells and whistles. In addition, the overall pipeline of our algorithm TransWCD-DL significantly exceeds the state-of-the-art method BGMix, with an F1 score improvement of +9.55\%, an OA improvement of +13.94\%, and an IoU improvement of +13.49\%, respectively. 

It is worth noting that BGMix is a background-mixed augmentation, and its performance is obtained by treating the models of other methods as baselines. Most of these models are designed for other dense prediction tasks. The suboptimal performance of BGMix can be attributed to the fact that the baseline models adopted by BGMix is less well-suited than our TransWCD for WSCD. This fact implies that an effective WSCD baseline, as the most fundamental part, is crucial for maximizing algorithmic performance potential. 

Furthermore, TransWCD surpasses our previous attempt with basic CAM, achieving a significant improvement of +14.13\% in F1 score, +25.94\% in precision, +2.17\% in recall, +17.88\% in OA, and +14.81\% in IoU. These results provide strong evidence for the necessity of constructing TransWCD, highlighting its superior performance compared to the basic CAM method.

We visualize the predicting results of TransWCD and TransWCD-DL compared to WSCD methods in Fig. \ref{experiment3}.  TransWCD-DL tackles the challenge of change fabricating commonly encountered in WSCD,  as illustrated in the 1st row. TransWCD-DL proficiently diminishes the problem of change missing, as evident in WCDNet and FCD-GAN (2nd row), as well as in PCA-Kmeans* \cite{kal2021} and BGMix (3rd row). 

Moreover, the following three rows demonstrate that Our TransWCD exhibits a significant improvement in capturing precise change regions and delineating boundaries compared to existing WSCD methods. Additionally, TransWCD-DL enhances the change predictions obtained by TransWCD. These qualitative evaluations provide compelling evidence for the outstanding efficacy of TransWCD and TransWCD-DL.

\subsubsection{Fully-Supervised Change Detection}
We further compare TransWCD and TransWCD-DL with the state-of-the-art FSCD methods on WHU-CD, DSIFN-CD, and LEVIR-CD datasets, manifesting the performance gap WSCD still needs to narrow.  The DSIFN-CD and LEVIR-CD datasets are commonly used in fully supervised settings but have not been adopted by WSCD methods for performance assessment. As shown in Table \ref{FSSS}, these  renowned FSCD methods are FC-EF \cite{47daudt2018fully}, FC-Siam-Di \cite{47daudt2018fully}, FC-Siam-Conc \cite{47daudt2018fully}, STANet \cite{34bandara2020spatial}, IFNet \cite{35zhang2020deeply}, DTCDSCN \cite{40liu2021building}, and SNUNet \cite{37fang2022snunet}.

Our partial predicting results closely match or even exceed several FSCD methods. On WHU-CD dataset, TransWCD-DL achieves comprehensive performance improvements over FC-EF \cite{47daudt2018fully}, FC-Siam-Di \cite{47daudt2018fully}, and FC-Siam-Conc \cite{47daudt2018fully}. Specifically, TransWCD-DL demonstrates gains of +2.58\%, +13.14\%, and +5.32\% in F1 score compared to these three FSCD methods.  On DSIFN-CD dataset, our TransWCD and TransWCD-DL surpass some FSCD methods in terms of Precision, Recall, and OA. Moreover, the gaps in F1 score between TransWCD-DL and all these state-of-the-art FSCD methods on DSIFN-CD  dataset are within 10\%. Additionally, Some FSCD methods cost significantly higher computational overhead. For example, SNUNet demonstrating the best FSCD performance, has FLOPs of 54.82G \cite{feng2023change}, while our TransWCD-DL is 13.27G. 

\begin{figure}
	\centering
	\includegraphics[scale=0.38]{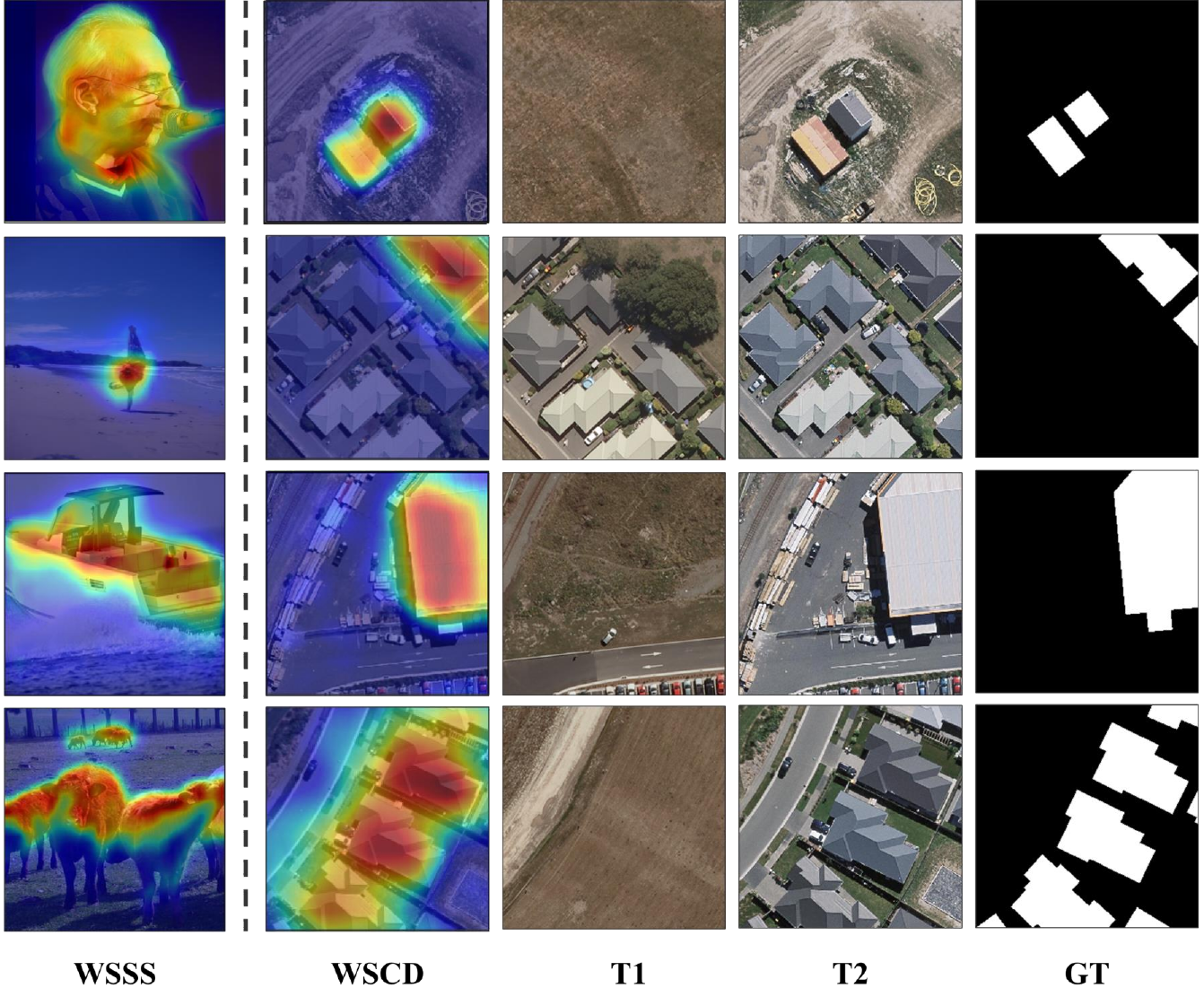}
	\caption{ CAM generated in weakly-supervised change detection (WSCD) exhibits more accurate predicted regions compared to weakly-supervised semantic segmentation (WSSS).}
	\label{discussion1}
\end{figure}

\section{Discussion}
\subsection{CAMs Give Much More to WCD than to WSSS}
\label{sec:cam}
Our TransWCD demonstrates remarkable performance in WSCD task by utilizing simple multi-scale CAMs, surpassing our initial expectations. This outstanding performance makes us question why stand-alone CAMs, which have yet to show comparable success in other dense prediction tasks like WSSS, perform well in WSCD. To investigate further, we visually examined the CAM density of WSSS and WSCD samples.

In Fig. \ref{discussion1}, we observe that for WSSS tasks, only certain regions are highlighted. For instance, these cows' heads and upper bodies are primarily attended to, while other body parts are disregarded. By contrast, in WSCD tasks, attention is given to the entire buildings, and the attended regions are relatively complete. We conduct some analyses about this.

It is well-known that CAMs are based on image-level classification tasks, focusing solely on discriminative features for classifying each category, while disregarding shared features among different categories. Hence, in WSSS, CAMs attend to discriminative features like the heads of cows while ignoring inconspicuous body features. However, such shared features may not exist in the regions of interest for WSCD. a) WSCD is a binary classification task with only two categories: changed or unchanged. b) The determination of changed or unchanged prediction relies on the presence of  coupled bi-temporal images, naturally incorporating region constraints.

Based on the above analysis, incomplete regions in WSCD may be non-problematic, and CAMs appear suitable for WSCD. In contrast, the issue of boundary refinement is a primary concern that demands our endeavor. 
\subsection{Does Transformer-Based WSCD Require Additional Shortcuts?}
\begin{figure}
	\centering
	\includegraphics[scale=0.5]{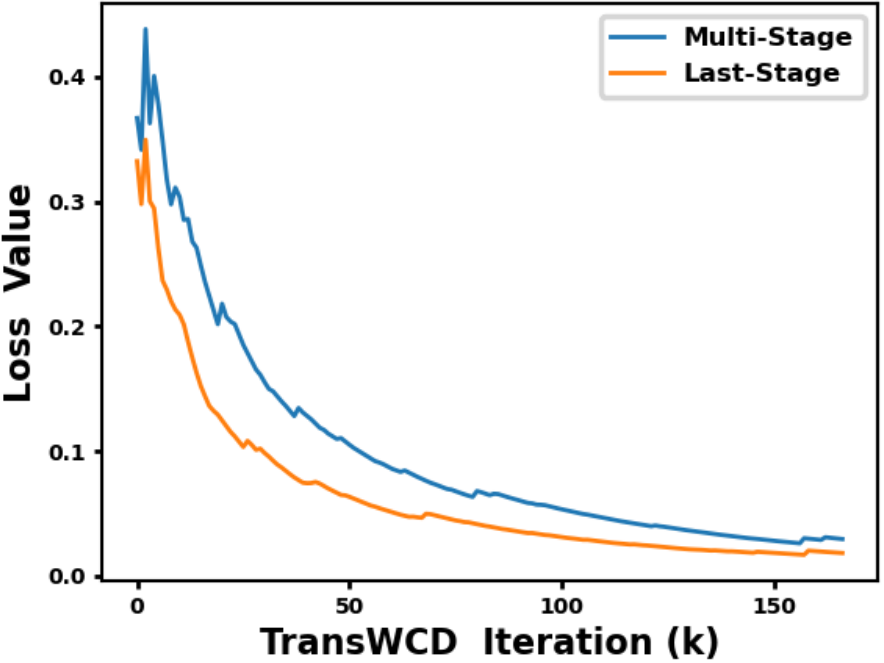}
	\caption{TransWCD with the last-stage CAMs converges faster than its multi-stage CAMs counterpart.}
	\label{discussion2}
\end{figure}

The construction of TransWCD is characterized by simplicity, devoid of any additional multilayer shortcuts, which adopted by some FSCD transformer-based methods \cite{bandara2022, Zhang2022}. In this section, we analyze the necessity of additinally shortcuts in transformer-based WSCD. Fig. \ref{discussion2} depicts loss reduction curves for last-stage CAMs with multiple scales and multi-stage CAMs. The last-stage CAMs model achieves faster convergence with approximately 10,000 fewer iterations than multi-stage CAMs. Furthermore, the last-stage CAMs yield smaller final loss values and a noteworthy 1.61\% improvement in F1 score compared to the multi-stage counterparts.

Regarding the results presented above, we speculate the following possibilities. 1) The multi-stage design increases model complexity, making it challenging to train effectively. 2) Feature granularity plays a role, where introducing coarse-grained features through the multi-stage design seems redundant and unnecessary for WSCD. WSCD typically deals with remote-sensing objects, where mostly are relatively small in scale. In comparison, the last-stage CAMs, with a more flexible scale zoom, could accurately handle diverse sizes of remote-sensing objects, including more fine-grained details. 

More importantly, the transformer architecture inherently incorporates shortcuts, rendering additional ones potentially superfluous. In the context of transformer-based WSCD, a more pragmatic approach to achieve multi-scale representation could entail employing scaling operations at the final stage of models.

\section{Conclusion}
This paper explores efficient approaches for weakly-supervised change detection (WSCD). We propose TransWCD, a powerful yet simple transformer-based model, which could serve as a baseline for future WSCD research. To tackle the dilemma of change missing and fabricating, we further develop TransWCD-DL by incorporating the global-scale and local-scale priors inherent in WSCD. TransWCD-DL consists of two additional components: a dilated prior (DP) decoder and a label-gated (LG) constraint. DP decoder is guided by the global-scale prior and selectively focuses on samples with the changed image-level label, while leveraging an all-unchanged pixel-level label as supervision for samples with the unchanged label. LG constraint acts as a penalty triggered when the translation from image-level labels to pixel-level predictions fails. With the guidance of these priors and the incorporation of these components, TransWCD-DL mitigates change missing and fabricating, and synchronously yields improved prediction performance. These priors and designs provide valuable insights for WSCD.

The experimental results demonstrate the effectiveness of TransWCD and TransWCD-DL. Moreover, we provide in-depth discussions about the factors that contribute to the algorithm's superior performance, highlighting the distinctive characteristics of the WSCD task and transformer-based WSCD model. Additionally, we argue that the challenge of predicting numerous small-scale changed targets within the WSCD paradigm, as exemplified in datasets like LEVIR-CD, warrants future research.

{\bibliographystyle{IEEEtran}
\bibliography{reference}
}

\end{document}